\def\year{2020}\relax
\documentclass[letterpaper]{article} 
\usepackage{aaai20}  
\usepackage{times}  
\usepackage{helvet} 
\usepackage{courier}  
\usepackage[hyphens]{url}  
\usepackage{graphicx} 
\usepackage{graphicx}  
\usepackage{amsmath}
\usepackage{amsfonts}
\usepackage{nicefrac}
\usepackage{color}

\DeclareMathOperator*{\argmin}{arg\,min}

\title{LagNetViP: A Lagrangian Neural Network for Video Prediction}
\author{Christine Allen-Blanchette, Sushant Veer, Anirudha Majumdar, Naomi Ehrich Leonard\\ 
Department of Mechanical and Aerospace Engineering,
Princeton University\\
\{ca15, sveer, ani.majumdar, naomi\}@princeton.edu
}
\begin{document}

\maketitle

\begin{abstract}
The dominant paradigms for video prediction rely on opaque transition models where neither the equations of motion nor the underlying physical quantities  of  the  system  are  easily  inferred. The  equations of motion, as defined by Newton’s second law, describe the time evolution of a physical system state and can therefore be applied toward the determination of future system states. In this paper, we introduce a video prediction model where the equations of motion are explicitly constructed from learned representations of the underlying physical quantities. To achieve this, we simultaneously learn a low-dimensional state representation and system Lagrangian. The kinetic and potential energy terms of the Lagrangian are distinctly modelled and the low-dimensional equations of motion are explicitly constructed using the Euler-Lagrange equations. We demonstrate the efficacy of this approach for video prediction on image sequences rendered in modified OpenAI gym \texttt{Pendulum-v0} and \texttt{Acrobot} environments. 

\end{abstract}

\section{Introduction}
\begin{figure}[t]
\begin{center}
\centerline{\includegraphics[width=.75\columnwidth]{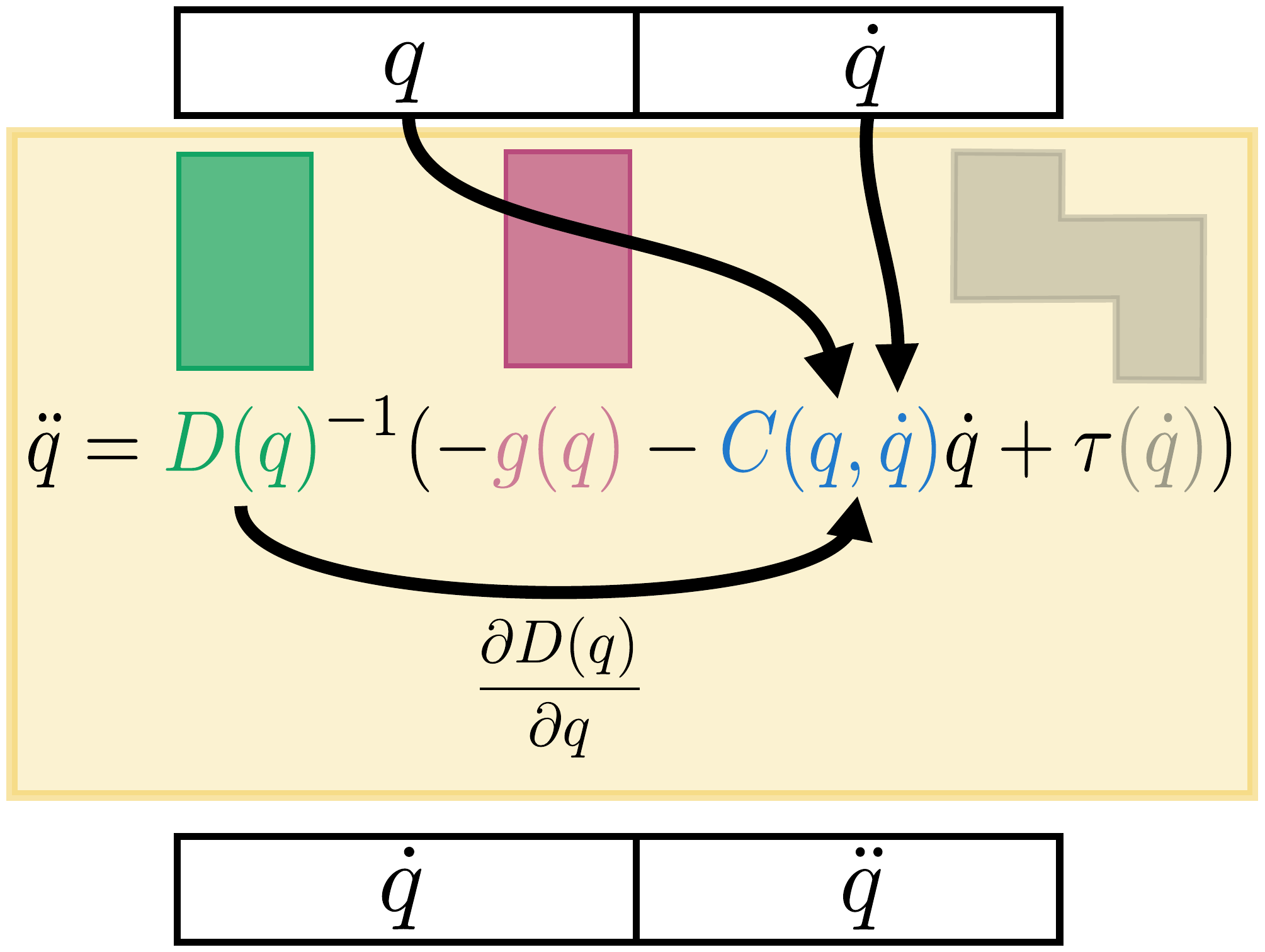}}
\caption{Lagrangian neural network for video  prediction (LagNetViP) architecture. LagNetViP (yellow) is comprised of an inertia tensor network (green), potential energy network (purple) and generalized forces network (gray). An auto-encoding network (not shown) can be trained end-to-end with LagNetViP to simultaneously learn a representation of the system state $(q, \dot{q})$ and the underlying dynamics. The network can also be adapted to settings where the generalized force $\tau$ is available as an input.}
\label{fig:architecture}
\end{center}
\vskip -0.2in
\end{figure}
Video prediction is a longstanding challenge in computer vision. While neural network models have provided significant advancements in the high-dimensional data regime, these advancements typically fail to provide an interpretable representation of the data-generating distribution. Juxtaposed against increased adoption of neural learning systems in safety critical domains, a growing segment of the community is advocating for the incorporation of physical priors to improve both the interpretability of representations and the generalizability of neural network models \cite{lake2017building,higgins2018towards}. In line with this call, we present a video prediction model that utilizes well-understood physical principles to capture the time evolution of the underlying data-generating process. In video prediction, high-dimensional images represent observations corresponding to low-dimensional states in a dynamical system. This correspondence is captured by the manifold hypothesis, which posits the existence of a low-dimensional manifold for high-dimensional data \cite{bengio2013representation}. By identifying the low-dimensional state-space manifold, well-understood physical principles can be leveraged to model the evolution of the system.

The time evolution of a physical system is determined by its equations of motion. 
Given the system Lagrangian, which is system kinetic energy minus system potential energy, the equations of motion can be derived using the Euler-Lagrange equations 
\cite{goldstein2002classical}. Previous works \cite{cranmer2020lagrangian,lutter2019deep} have shown that it is possible to learn the system Lagrangian from low-dimensional measurements. In this work, we present an approach to simultaneously learn a low-dimensional state-space representation and system Lagrangian from high-dimensional image data. In our approach, images are mapped to and from the low-dimensional state-space by an auto-encoding neural network and initial states are integrated forward using the equations of motion determined by the system Lagrangian and Euler-Lagrange equations. Towards interpretability of the representation, 
the inertia tensor, which determines the kinetic energy, and the potential energy 
are parameterized by distinct neural networks. A pictorial representation of our model (LagNetViP) is given in Figure \ref{fig:architecture}. 

The efficacy of our approach requires that the auto-encoder and system Lagrangian agree on an appropriate low-dimensional representation. To encourage this, we define the training loss as the sum of three terms: (1) the reconstruction loss of the auto-encoded image sequence, (2) the reconstruction loss of the latent trajectory generated by the learned dynamics and (3) a mean absolute difference between the auto-encoded image sequence and the latent trajectory generated by the learned dynamics. We validate the importance of each term through an ablation study and provide a qualitative assessment using image sequences rendered in modified OpenAI gym \texttt{Pendulum-V0} and \texttt{Acrobot} environments.

\section{Related work}
Recurrent neural networks (RNNs) have been applied to a broad class of sequence prediction problems including language modeling, machine translation, image processing and audio processing \cite{graves2013generating,cho2014learning,srivastava2015highway,oord2016wavenet}. The utility of RNNs stems from the architectural structure, which enforces parameter sharing and encourages the network to learn statistics that generalize across sequences.

When sequential data can be interpreted as the observations of a dynamical system, the structure of difference or differential equations can be leveraged for prediction. Sequential image data is necessarily discrete although the generating process for image data is continuous. Consequently, there is a disparity in modeling approaches; \citeauthor{watter2015embed} and \citeauthor{karl2016deep} model the nature of the data with a (generative) discrete-time dynamical model, while \citeauthor{yildiz2019ode} model the continuous generating process with a continuous-time dynamical model. The related work of \cite{chen2018neural} facilitates estimation of continuous-time dynamics with neural networks by allowing for backpropagation through arbitrary ODE solvers.

The structure of the output space of a dynamical system can also be leveraged in learning \cite{stewart2017label}.
In \cite{greydanus2019hamiltonian,toth2019hamiltonian,bertalan2019learning}, the authors model the underlying dynamics of sequential data assuming the Hamiltonian structure. Dynamical state updates are generated using Hamilton's equations. For high-dimensional data, \citeauthor{greydanus2019hamiltonian} and \citeauthor{toth2019hamiltonian} learn a map from the observation space to a low-dimensional phase space where Hamilton's equations can be applied.
A limitation of these approaches is that they are only applicable in the context of conservative systems. In \cite{desmond2019symplectic}, the authors surpass this limitation by assuming the Port-Hamiltonian structure, which can be used to model nonconservative systems. 

In \cite{lutter2019deep,cranmer2020lagrangian,saemundsson2019variational,zhong2020unsupervised}, the authors assume the Lagrangian structure which allows for incorporation of nonconservative forces naturally. \citeauthor{saemundsson2019variational} and \citeauthor{zhong2020unsupervised} introduce generative models that leverage the variational auto-encoder (VAE) formulation to learn a representation of the low-dimensional state-space from high-dimensional images.  \citeauthor{saemundsson2019variational} use a Gaussian prior on the latent code but set its dimension substantially higher than the number of degrees of freedom inherent to the system. This architectural choice limits interpretability of the learned coordinate representation and may have adverse effects in the control setting.
In \cite{zhong2020unsupervised} (work concurrent to ours), the authors select the latent dimension according to the number of degrees of freedom in the system. They find that with this choice the standard Gaussian prior severely inhibits learning and instead choose to apply system specific priors on the latent code.

In this work we introduce a discriminative model with latent dimension consistent with the number of degrees of freedom inherent to the system 
but without the need for system specific structural priors. Other discriminative models that leverage the Lagrangian structure are \cite{lutter2019deep} and \cite{cranmer2020lagrangian}. Our model differs from \cite{lutter2019deep} in that we learn and apply the \emph{forward} model for dynamical state prediction whereas \cite{lutter2019deep} learn the \emph{inverse} model for generalized force prediction and control. Moreover, we apply our approach to high-dimensional observations which neither \cite{lutter2019deep} nor \cite{cranmer2020lagrangian} attempt.

\section{Lagrangian dynamics} 
\label{sec:lagrangian-dynamics}
In this section, we provide a brief introduction to Lagrangian dynamics. For the sake of brevity, the exposition is kept terse; interested readers can find further details in 
\cite{spong2008robot,goldstein2002classical}. 

A Lagrangian mechanical system has an associated configuration space $\mathcal{Q}$ which, loosely speaking, includes all the feasible configurations (or poses) $q$ of the mechanical system; e.g., for a simple pendulum, the configuration space can be defined as the space $\mathcal{Q}:=[-\pi, \pi)$ of all possible angles. We denote the rate of change of the configuration $q$ by $\dot{q}$. Mathematically, $\mathcal{Q}$ takes on the structure of a manifold with $q\in\mathcal{Q}$ being a set of coordinates on it and the tuple $(q,\dot{q})$ lies in its tangent bundle $\mathcal{TQ}$. The dimension of $\mathcal{Q}$ corresponds to the degrees of freedom of the dynamical system and will be denoted by $m\in\mathbb{N}$ throughout the paper.

The Lagrangian $L:\mathcal{TQ}\to\mathbb{R}$ is a function that maps the tangent space $\mathcal{TQ}$ to a scalar. To define the Lagrangian first requires the introduction of the kinetic energy $T:\mathcal{TQ}\to [0,\infty)$ and the potential energy $V:\mathcal{Q}\to\mathbb{R}$. Intuitively, the kinetic energy $T$ is the energy possessed by a mechanical system by virtue of its motion whereas the potential energy $V$ is the energy stored in a mechanical system due to its configuration. We assume the form of the kinetic energy to be quadratic in the velocity, as follows: 
\begin{equation}
\label{eq:kinetic}
T = \frac{1}{2} \dot{q}^{\rm T}\, D(q)\, \dot{q}
\end{equation}
where $D:\mathcal{Q}\to \mathbb{S}^n_{++}$ is the positive definite inertia matrix. The kinetic energy of any mechanical system with holonomic constraints satisfies the quadratic form \eqref{eq:kinetic} --- see \cite{spong2008robot}
for examples. 
Hence, learning the quadratic form \eqref{eq:kinetic} allows us to embed more structure in our architecture without compromising the richness of the class of systems that can be addressed by our approach. Indeed, the quadratic form of the kinetic energy has also been adopted in SymODEN \cite{desmond2019symplectic} and DeLaN \cite{lutter2019deep}.
With the kinetic and the potential energy introduced above, we can express the Lagrangian $L$ as the difference between them:
\begin{equation}
\label{eq:Lagrangian}
L(q,\dot{q}) = T(q,\dot{q}) - V(q).
\end{equation}

The equations of motion for the Lagrangian dynamical system can be conveniently uncovered by the following operation on the Lagrangian (Euler-Lagrange equations):
\begin{equation}
\label{eq:eom}
\frac{d}{dt}\frac{\partial L}{\partial \dot{q}} - \frac{\partial L}{\partial q} = \tau
\end{equation}
where $\tau$ represents the generalized forces, which encapsulate the effect of exogenous influences on the evolution of the Lagrangian system. For systems for which the total energy is conserved, $\tau=0$. However, 
the explicit presence of $\tau$ in our Lagrangian formalism allows us to handle  non-conservative influences with ease, e.g., friction in the pendulum system. 

The equations of motion \eqref{eq:eom} can be expanded further by using \eqref{eq:kinetic} and \eqref{eq:Lagrangian} in \eqref{eq:eom} giving:
\begin{equation}
\label{eq:eom-tau}
D(q)\ddot{q} + C(q,\dot{q}) + g(q) = \tau \enspace,
\end{equation}
where $D(q)$ is the inertia matrix, $C(q,\dot{q})$ is the Coriolis term, $g(q):=\frac{\partial V(q)}{\partial q}$, and $\tau$ are the generalized forces.

Noting that $D(q)$ is positive definite, we can solve for the acceleration $\ddot{q}$:
\begin{equation}
\label{eq:eom-rayleigh}
\ddot{q} =  -D(q)^{-1}\Big( C(q,\dot{q}) + g(q) - \tau \Big) \enspace.
\end{equation}

The Lagrangian dynamics admit further structure that allows us to express the components of the Coriolis term $C$ as a function of the components of the inertia matrix $D$. For any $k\in\{1,2,\cdots,m\}$, we can express the the $k$-th row of $C$, denoted by $C_k(q,\dot{q})$, as $C_k(q,\dot{q}) = \sum_{i,j} c_{ijk} \dot{q}_i\dot{q}_j$. Further,   
the terms $c_{ijk}$ take the form \cite{spong2008robot}:
\begin{equation*}
    c_{ijk} = \left\{ \frac{\partial d_{k,j}}{\partial q_i} - \frac{1}{2}\frac{\partial d_{ij}}{\partial q_k} \right\} \enspace.
\end{equation*}

\section{Lagrangian neural networks}
\begin{figure*}[!ht]
\includegraphics{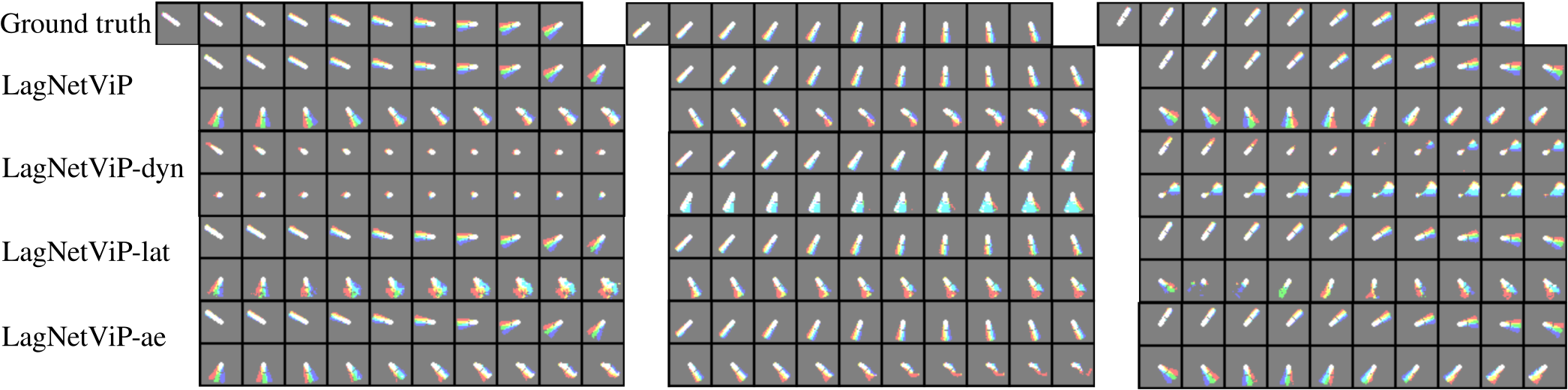}
\caption{Ablative analysis of the LagNetViP loss function on the pendulum dataset. Top to bottom. (1) Three randomly selected testset trajectories, (2) LagNetViP generated trajectory (two rows), (3) LagNetViP-dyn generated trajectory (two rows), (4) LagNetViP-lat generated trajectory (two rows), (5) LagNetViP-ae generated trajectory (two rows). The testset trajectories for the pendulum dataset consist of 10 observations (over time). To demonstrate how well the dynamics are learned, the first observation of each testset trajectory is encoded to the low-dimensional state-space and integrated forward 20 steps (over time) according to the learned dynamics model. This trajectory is reconstructed giving the image sequences show in the figure. Consider LagNetViP in the left most panel. The first row (frames 1-10) illustrates how well the model extrapolates within the range of the testset and the second row (frames 11-20) illustrates how well the model can extrapolate (forward in time) beyond the range of the testset.  
}
\label{fig:pendulum}
\end{figure*}
In this section, we review estimation of the system Lagrangian in the context of low-dimensional state-space measurements and introduce a strategy for estimating the system Lagrangian from (high-dimensional) image data.

\subsection{Learning from state-space measurements}
\label{sec:LagNet}
We outline first our approach for estimating the system Lagrangian from low-dimensional position-velocity measurements. The system Lagrangian is modelled as the difference between the kinetic and potential energies where the mass matrix $D_\theta$ and potential energy function $V_\phi$ are parameterized by neural networks
\begin{equation*}
L(q,\dot{q}) = \frac{1}{2}\dot{q}^T D_\theta(q) \,\dot{q} - V_\phi(q).
\end{equation*}
This formulation differs from previous work~\cite{cranmer2020lagrangian} where 
neither the kinetic energy nor the potential energy are explicitly modelled and from~\cite{lutter2019deep} where the potential force is modelled instead of the potential energy. To ensure invertibility of the mass matrix $D_\theta$ we use an intermediate matrix $J_\theta$ of the same shape and compute $D_\theta$ by
\begin{equation*}
    D_\theta(q) = J_\theta(q)^T J_\theta(q) + \lambda I.
\end{equation*}
In our experiments $\lambda$ is set to the dimension of the position vector $q$.

The parameters $(\theta,\phi)$ of the system Lagrangian are estimated by iterative minimization of the mean absolute difference between ground truth position-velocity measurement sequences denoted $Z_n=\{z^n_1,\dots, z^n_T\}$, with $z^n_t = (q^n_t, \dot{q}^n_t)$ and $n\in\{1,\dots,N\}$, and predicted sequences $\hat{Z}_n$ defined similarly. 
The minimization problem is given by
\begin{equation}
    (\theta^*, \phi^*) = \argmin_{\theta,\phi} \frac{1}{N}\sum_{n,t} |z_t^n - \hat{z}_t^n|\quad t\in\{1,\dots,T\}.
\end{equation}
Predicted sequences are computed recursively from an initial ground truth measurement $z^n_0$ and the current values of $\theta$ and $\phi$.
Concretely, the predicted measurement $\hat{z}^n_{t+1}$ is computed from the preceding measurement $\hat{z}^n_{t}$ with $\hat{z}^0_n := z_n^0$
and the dynamical update $\dot{z}^n_t$ given by 
\begin{equation*}
    \dot{z}^n_t = \left(\dot{q}^n_t\,,\;\;
    -D_\theta(q^n_t)^{-1}\Big( C_\theta(q^n_t,\dot{q}^n_t) + g_\phi(q^n_t) - \tau\Big)\right).
\end{equation*}
The Coriolis term $C_\theta$ and potential force $g_\phi$ are discussed in the previous section.
In our experiments we consider systems without external forces, that is $\tau=0$, and perform numerical integration with the Euler method. Note that the use of more sophisticated numerical integrators, in particular variational integrators, would likely improve performance.
\subsection{Learning from image sequences}
In this section we outline our approach for estimating the state-space representation and system Lagrangian from high-dimensional observations\footnote{Since we cannot perform velocity prediction from a single image we use observations which we define as image tuples.}. 
Our prediction pipeline maps high-dimensional observations
to a low-dimensional state-space representation whose structure is learned during the training phase. We predict future low-dimensional states using the Euler-Lagrange equations, then map the resulting sequence back to the observation space giving the predicted image sequence.

To compress and reconstruct observations of system trajectories given as temporal sequences of $T$ high-dimensional observations $X_n =\{x^n_0, \dots, x^n_T\}$, with $n\in\{0,\dots,N\}$, we use an auto-encoding neural network $f=D_\varphi\circ E_\phi$, where $D_\varphi$ 
and $E_\phi$ are decoding and encoding networks respectively.
We denote encoded observation sequences $Z_n=\{z^n_1,\dots, z^n_T\}$ and interpret $z^n_t = E_\phi(x^n_t)$ as a position-velocity measurement. Predicted sequences $\hat{Z}_n=\{\hat{z}^n_1,\dots, \hat{z}^n_T\}$ are determined as described in the previous section.

The parameters $(\varphi,\phi)$ of the auto-encoding network and $(\theta,\psi)$ of the system Lagrangian are estimated jointly by iterative minimization of the three component loss function:
\begin{equation*}
\mathcal{L}_{total} = \mathcal{L}_{ae} + \mathcal{L}_{dyn} + \gamma \,\mathcal{L}_{lat},
\end{equation*}
where $\mathcal{L}_{ae}$ is the auto-encoding reconstruction loss:
\begin{equation*}
    \mathcal{L}_{ae}(\pmb{X}) = \frac{1}{NT}\sum_{n=0}^{N}\sum_{t=0}^{T-1}\|x_t^n - f\left(x_t^n\right)\|^2_2,
\end{equation*}
with $\pmb{X}=\{X_0, \dots, X_{N}\}$; $\mathcal{L}_{dyn}$ is the predicted sequence reconstruction loss:
\begin{equation*}
    \mathcal{L}_{dyn}(\pmb{X}) = \frac{1}{NT}\sum_{n=0}^{N}\sum_{t=1}^T\|D_\varphi(\hat{z}_t^n) - x_t^n\|^2_2;
\end{equation*}
and $\mathcal{L}_{lat}$ is the distance between encoded and predicted sequences:
\begin{equation*}
    \mathcal{L}_{lat}(\pmb{X}) = \frac{1}{N}\sum_{n=1}^{N}\sum_{t=0}^T\|\hat{z}_t^n - E_\phi\left(x_t^n\right)\|^2_2.
\end{equation*}
This formulation differs from previous work~\cite{greydanus2019hamiltonian} where a structural prior is imposed on the embedding space. In our experiments we set $\gamma=0.1$. 

\section{Empirical analysis}
In this section we empirically validate the proposed approach. Toward this end, we perform an ablation study where components of the cost function are removed and a qualitative comparison of the results are presented. Specifically, we compare the following models:
\begin{enumerate}
\item LagNetViP: The proposed model that simultaneously learns a low-dimensional state-space representation and system Lagrangian from high-dimensional image data. The training loss is the sum of three terms: (1) the auto-encoder reconstruction loss, (2) the reconstruction loss of the latent trajectory generated by the learned dynamics and (3) a mean absolute difference between the auto-encoded image sequence and the latent trajectory generated by the learned dynamics. 
\item LagNetViP-dyn: The proposed model trained without a reconstruction loss on the latent trajectory generated by the learned dynamics.
\item LagNetViP-lat: The proposed model trained without a mean absolute difference between the auto-encoded image sequence and the latent trajectory generated by the learned dynamics.
\item LagNetViP-ae: The proposed model trained without the auto-encoder reconstruction loss.
\end{enumerate}

In each model, we use a symmetric auto-encoding network to map high-dimensional observations to a low-dimensional state-space representation. The encoding network is a four layer neural network: three convolutional layers followed by a fully-connected layer (see Table \ref{encoder-configuration}); where convolutional layer is followed by a \texttt{ReLU} nonlinear unit. Both the inertia tensor and potential energy functions are parameterized by three layer fully-connected neural networks with 200 hidden units in each layer and \texttt{tanh} nonlinear units. 

\begin{table}[t]
\caption{Encoder configuration. Filter column: (Layers 1-3) width, height, channel and output dimension of convolutional filters; (Layer 4P) the input and output dimension of the fully-connected layer for the pendulum experiment; (Layer 4A) the input and output dimension of the fully-connected layer for the Acrobot experiment.} \smallskip
\centering
\smallskip\begin{tabular}{|c|c|c|c|c|}
\hline
Layer & Filter & Stride & Padding\\
\hline
1 & 4x4x3x12 & 2 & 1\\
2 & 4x4x12x24 & 2 & 1\\
3 & 4x4x24x12 & 2 & 1\\
4P & (4*4*12)x4 & 2 & 1\\
4A & (4*4*12)x6 & 2 & 1\\
\hline
\end{tabular}
\label{encoder-configuration}
\end{table}

To demonstrate the efficacy of our approach on video prediction problems, we consider image sequences generated using modified OpenAI gym \texttt{Pendulum-V0} and \texttt{Acrobot} environments. To generate image sequences we modify the OpenAI gym environments to use RK4 integration instead of Euler integration and increase the width of the pendulum and Acrobot arms in rendering. 

\paragraph{Pendulum Dataset.}
The pendulum dataset consists of $N=10,000$ trajectories. Each trajectory is comprised of $T=10$ observations and each observation is constructed by concatenating three sequential images along the channel dimension. We train on 8,000 of the 10,000 trajectories using the Adam optimizer \cite{kingma2014adam} with a learning rate of $1\times10^{-3}$ and weight decay of $1\times10^{-5}$.

\paragraph{Acrobot.}
The Acrobot dataset consists of $N=10,000$ trajectories. Each trajectory is comprised of $T=2$ observations and each observation is constructed by concatenating three sequential images along the channel dimension. We train on 8,000 of the 10,000 trajectories using the Adam optimizer \cite{kingma2014adam} with a learning rate of $1\times10^{-4}$ and weight decay of $1\times10^{-5}$.

\subsection{Results}
\begin{figure}[!ht]
\includegraphics{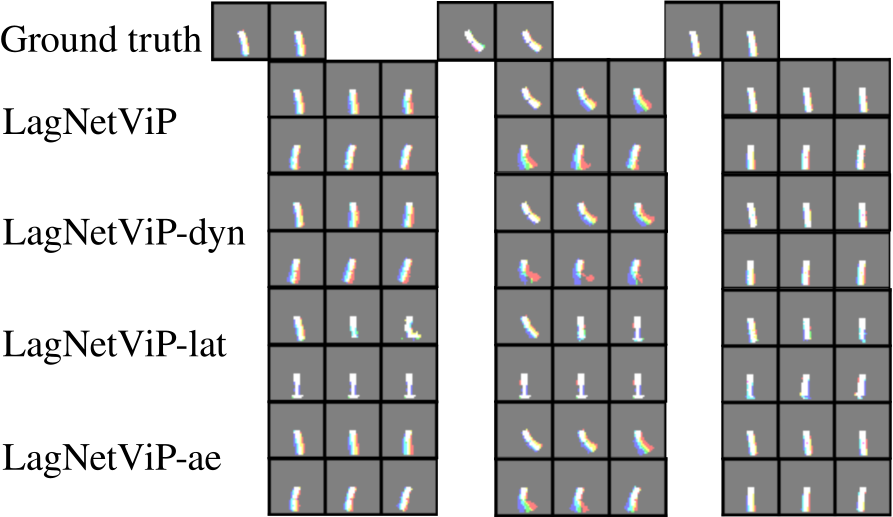}
\caption{Ablative analysis of the LagNetViP loss function on the Acrobot dataset. Top to bottom. (1) Three randomly selected testset trajectories, (2) LagNetViP generated trajectory (two rows), (3) LagNetViP-dyn generated trajectory (two rows), (4) LagNetViP-lat generated trajectory (two rows), (5) LagNetViP-ae generated trajectory (two rows). The testset trajectories for the pendulum dataset consist of 2 observations (over time). To demonstrate how well the dynamics are learned, the first observation of each testset trajectory is encoded to the low-dimensional state-space and integrated forward 6 steps (over time) according to the learned dynamics model. This trajectory is reconstructed giving the image sequences show in the figure. Consider LagNetViP in the left most panel. The first row (frames 1-3) illustrates how well the model extrapolates within the range of the testset and the second row (frames 4-6) illustrates how well the model can extrapolate (forward in time) beyond the range of the testset. 
}
\label{fig:acrobot}
\end{figure}
Figures \ref{fig:pendulum} and \ref{fig:acrobot} present qualitative comparisons on randomly selected testset trajectories from the pendulum and Acrobot datasets. On both datasets LagNetViP is able to extrapolate beyond the last observation in the testset trajectory. Removing the reconstruction loss on the latent trajectory generated by the learned dynamics has the most disastrous effect on reconstruction quality. This is evident in the fact that LagNetViP-dyn is unable to provide reasonable reconstructions even within the range of the testset in some cases. LagNetViP-lat performs well within the testset range but is unable to extrapolate farther and LagNetViP-ae exhibits good performance but fails to out perform LagNetViP. 

Consider the trajectory generated by the LagNetViP model in the left-most panel of Figure \ref{fig:pendulum}. The position of the pendulum begins pointing up and to the left of the fixed point. In the first 10 frames, the pendulum swings down but not quite reaching the downward pointing position, as in the trajectory of the testset. LagNetViP is able to extrapolate beyond the testset trajectory predicting the upward swing of the pendulum on the right side through the downward pointing position.

\section{Discussion and Conclusion}
In this work, we introduce a video prediction model where equations of motion are explicitly constructed from learned representations of underlying physical quantities. The low-dimensional state-space representation and system Lagrangian are learned simultaneously. Images are mapped to and from the low-dimensional state-space by an auto-encoding network and initial states are integrated forward using the equations of motion determined by the system Lagrangian and Euler-Lagrange equations. The approach excels over the baseline model with strong reconstruction performance on the pendulum system and strong indications of possibility on the chaotic Acrobot system.

\section{Acknowledgements}
We thank Shinkyu Park, Desmond Zhong, David Isele and Patricia Posey for their insights. This research has been supported in part by ONR grant \#N00014-18-1-2873 and by the School of Engineering and Applied Science at Princeton University through the generosity of William Addy~’82. 

\bibliography{example_paper.bib}
\bibliographystyle{aaai}

\end{document}